\documentclass[10pt, a4paper]{article}
\usepackage{lrec-coling2024}
\usepackage{bm}
\usepackage{color}
\usepackage{adjustbox}
\usepackage[most]{tcolorbox}
\usepackage{algorithm}
\usepackage{algorithmicx}
\usepackage{algpseudocode}
\usepackage{graphicx}
\usepackage{booktabs}
\usepackage{colortbl}
\usepackage{caption}
\usepackage{subcaption}
\usepackage{multicol}
\usepackage{enumitem}

\setlist[itemize]{noitemsep, nolistsep, leftmargin=*}
\setlist[enumerate]{noitemsep, nolistsep, leftmargin=*}

\algnewcommand\algorithmicinput{\textbf{Input:}}
\algnewcommand\algorithmicoutput{\textbf{Output:}}
\algnewcommand\algorithmicinitialize{\textbf{Initialize:}}
\algnewcommand\INPUT{\item[\algorithmicinput]}
\algnewcommand\OUTPUT{\item[\algorithmicoutput]}
\algnewcommand\INITIALIZE{\item[\algorithmicinitialize]}
\algnewcommand{\LineComment}[1]{\State \(\triangleright\) #1}

\definecolor{SRC}{HTML}{EF7A6D} % 239, 122, 109
\definecolor{TGT}{HTML}{9DC3E7} % 157, 195, 231
\definecolor{TEP}{HTML}{F1D77E} % 241, 215, 126
\definecolor{GRY}{HTML}{D8DEE9} % 216, 222, 233
\tcbset{
  on line, 
  boxsep=4pt, left=0pt,right=0pt,top=0pt,bottom=0pt,
  colframe=white,colback=SRC,  
  highlight math style={enhanced}
}

\newcommand{\ours}{\textsc{TamePT}}

\title{A Two-Stage Framework with Self-Supervised Distillation for Cross-Domain Text Classification}

\name{
  Yunlong Feng, Bohan Li, Libo Qin, Xiao Xu, Wanxiang Che\sthanks{~~Corresponding author.}
}

\address{
Research Center for Social Computing and Information Retrieval \\
Harbin Institute of Technology, China \\ 
\{ylfeng,bhli,lbqin,xxu,car\}@ir.hit.edu.cn
}

\abstract{
    Cross-domain text classification is a crucial task as it enables models to adapt to a target domain that lacks labeled data. 
    It leverages or reuses rich labeled data from the different but related source domain(s) and unlabeled data from the target domain.
    To this end, previous work focuses on either extracting domain-invariant features or task-agnostic features, ignoring domain-aware features that may be present in the target domain and could be useful for the downstream task.
    In this paper, we propose a two-stage framework for cross-domain text classification. 
    In the first stage, we finetune the model with mask language modeling (MLM) to learn from the source domain.
    In the second stage, we further fine-tune the model with \textit{self-supervised distillation} (SSD) and unlabeled data to adapt to the target domain.
    We evaluate its performance on a public cross-domain text classification benchmark and the experiment results show that our method achieves new state-of-the-art results for both single-source domain adaptations (94.17\% \begin{small}$\uparrow$1.03\%\end{small}) and multi-source domain adaptations (95.09\% \begin{small}$\uparrow$1.34\%\end{small}).
 \\ \newline \Keywords{Cross-domain text classification, Unsupervised domain adaptation}}

\begin{document}

\maketitleabstract

\section{Introduction}\label{sec:intro}

In the era of large models, neural network models have achieved remarkable results in a myriad of tasks. However, a prevalent challenge arises when these models, often trained on source domains, are deployed in different, target domains, leading to a domain shift~\cite{gretton2006kernel}.
Unsupervised Domain Adaptation~(UDA) emerges as a vital solution by aiming to adapt the models trained on source domains with labeled data to a target domain laden with unlabeled data. The significance of UDA is pronounced in the age of large models, which, despite their prowess, frequently require abundant labeled data for fine-tuning to attain optimal performance. By leveraging labeled data from source domains, UDA substantially mitigates this dependency, thereby eliminating the need for expensive and time-consuming annotation processes in the target domain. In this light, our paper delves into the subdomain of UDA, specifically focusing on cross-domain text classification.

\begin{figure*}[t]
  \centering

  \begin{subfigure}[b]{0.495\linewidth}
    \centering
    \includegraphics[page=2,height=3cm]{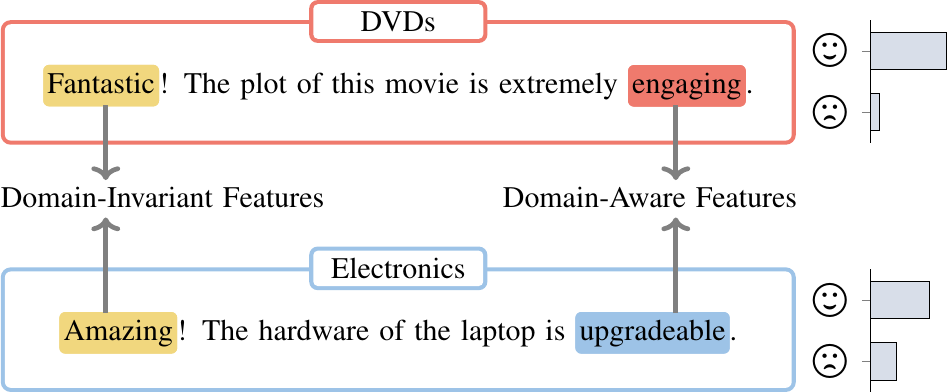}
    \caption{The model's predictions of DVDs (source domain) and Electronics (target domain) without adaptation.}
    \label{intro:a}
  \end{subfigure}
  \hfill
  \begin{subfigure}[b]{0.495\linewidth}
    \centering
    \includegraphics[page=3,height=3cm]{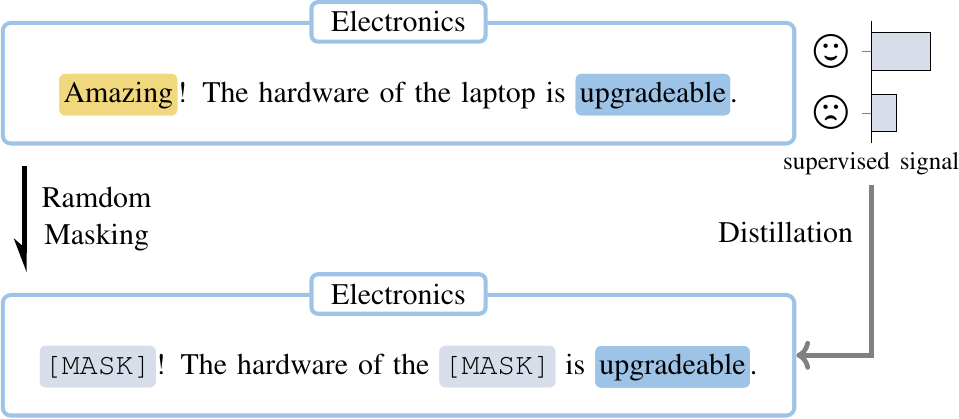}
    \caption{An overview of the self-supervised signal we constructed to guide model.}
    \label{intro:b}
  \end{subfigure}
  \caption{
    % Overview
    The colors mean domain-invariant (\tcbox[on line,boxsep=0pt,left=3pt,right=3pt,top=6pt,bottom=0pt,boxrule=0pt,colframe=white,colback=TEP]{}), source domain aware (\tcbox[on line,boxsep=0pt,left=3pt,right=3pt,top=6pt,bottom=0pt,boxrule=0pt,colframe=white,colback=SRC]{}) and target domain aware (\tcbox[on line,boxsep=0pt,left=3pt,right=3pt,top=6pt,bottom=0pt,boxrule=0pt,colframe=white,colback=TGT]{}).
    (a) The model can exploit the domain-invariant features but lacks the use of domain-aware features when predicting the target domain.
    (b) The supervised signal we construct, is designed to force the model to make a connection between predictions and latent domain-aware features of the target domain.
  }
  \label{intro}
\end{figure*}

Cross-domain text classification is encumbered by domain discrepancy emanating from variations in expressions across different domains. Addressing this conundrum, a substantial body of work \cite{clinchant-etal-2016-domain,10.1162/tacl_a_00328,zhou-etal-2020-sentix,du-etal-2020-adversarial,wu-shi-2022-adversarial} has been dedicated to extracting domain-invariant features between domains to bolster classification models' performance across multiple domains. Concurrently, another strand of work \cite {du-etal-2020-adversarial, karouzos-etal-2021-udalm} explores the utilization of language modeling to aid models in harnessing task-agnostic features in the target domain, thus enhancing their performance in cross-domain text classification tasks.

In spite of these advancements, not all features conducive to a given task exhibit domain-invariance, as illustrated in Figure \ref{intro:a}. For instance, while expressions like ``fantastic'' and ``amazing'' are domain-invariant and can convey positive sentiments universally, terms such as ``upgradeable'' pertain to specific contexts like electronic products but not to DVDs, representing what we term as domain-aware features. The exploration of such features and their relation to the task at hand is often overlooked by existing methods, leaving a gap in addressing domain-aware features in the target domain.

To bridge this gap, Figure \ref{intro:b} outlines our proposed approach which ingeniously constructs a self-supervised signal, enabling models to attend to the latent domain-aware features of the target domain. This is pivotal for large models, which often grapple with new data from domains different from their training corpus. By masking domain-invariant features, our approach forces the model to establish a correlation between the predictions and the remaining domain-aware features. Subsequently, the model reinforces this relationship when domain-aware features are masked, allowing it to focus on latent domain-aware features in the target domain through a process that we denote as \textit{self-supervised distillation}.

In this paper, we propose a novel cross-domain text classification model comprising a two-stage learning procedure: \textit{(1) learning from the source domain} and \textit{(2) adapting to the target domain}. This two-stage learning procedure significantly augments the model's performance and stability, rendering it a promising approach for cross-domain text classification tasks. Our experiment results on the Amazon reviews benchmark \citetlanguageresource{blitzer-etal-2007-biographies} substantiate that our proposed method sets new state-of-the-art results for both single-source domain adaptations (94.17\%\begin{small}$\uparrow$1.03\%\end{small}) and multi-source domain adaptations (95.09\%\begin{small}$\uparrow$1.34\%\end{small}). Furthermore, a detailed analysis accentuates the generalization and effectiveness of our method, heralding a significant stride in the realm of UDA and cross-domain text classification.

To summarize, our contributions are as follows:

\begin{itemize}
  \item We introduce \textit{self-supervised distillation}, a simple yet effective method that helps models better capture domain-aware features from unlabeled data in the target domain.
  \item We propose a two-stage learning procedure that enables existing classification models to adapt to the target domain effectively.
  \item The experiments on the Amazon reviews benchmark for cross-domain classification show that our proposed model achieves new state-of-the-art results.
\end{itemize}

\section{Background}

\subsection{Problem Formulation}

To establish basic notations for our study, we define a domain $\mathcal{D} = \{ \mathcal{X}, P(\mathcal{X})\} $, where $\mathcal{X}$ represents the input feature space (e.g., the text representations), and $P(\mathcal{X})$ denotes the marginal probability distribution over that feature space.
Let $\mathcal{T}$ define a task (e.g., sentiment classification) as $\mathcal{T} = \{ \mathcal{Y}, P (Y |X) \}$, where $\mathcal{Y}$ is the label space.
Moreover, a dataset is denoted by $\mathcal{D}^{\mathcal{T}} = \{ (\bm{x}_i, y_i)\}_{i=1}^n$, where $\bm{x}_i \in \mathcal{D}$ and $y_i \in \mathcal{Y}$.

In this paper, we focus on cross-domain text classification, which is a subdomain of UDA \cite{ramponi-plank-2020-neural}.
Specifically, we aim to learn a function $\mathcal{F}$ trained with labeled dataset $\mathcal{D}_{S}^{\mathcal{T}}$ and unlabeled dataset $\mathcal{D}_T$, which can effectively perform the task $\mathcal{T}$ in the domain $\mathcal{D}_T$.
Here we respectively denote $S$ and $T$ as the source domain and the target domain, and $P_S (\mathcal{X}) \neq P_T (\mathcal{X})$.

\subsection{Prompt Tuning}\label{prompt-tuning}
We use prompt tuning as the formula for the text classification task, which reformulates the downstream task into cloze questions through a textual prompt $\bm{x}_{\text{p}}$ \cite{petroni-etal-2019-language, brown2020language}.
Specifically, a textual prompt consists of an input sentence, a template containing \texttt{[MASK]}, and two special tokens (\texttt{[CLS]} and \texttt{[SEP]}):
\begin{equation}\label{textual-prompt}
  \bm{x}_{\text{p}} = ``\texttt{[CLS]}~\bm{x}\text{. It is }\texttt{[MASK]}\text{. }\texttt{[SEP]}",
\end{equation}
where $\bm{x}$ is the input sentence.

The Pretrained Language Model(PLM) takes the textual prompt $\bm{x}_{\text{p}}$ as input and utilizes contextual information to fill in the \texttt{[MASK]} token with a word from the vocabulary as the output.
The output word is subsequently mapped to a label $\mathcal{Y}$. Following \citet{wu-shi-2022-adversarial}, we use ``\{good,bad\}'' as the label words.
Finally, given an labeled dataset $\mathcal{D}^{\mathcal{T}} = \{ (\bm{x}_i, y_i)\}_{i=1}^n$, the PLM is finetuned by minimizing the cross-entropy loss.
The objective of prompt tuning can be defined as the following formula:
\begin{equation}
  \mathcal{L}_{pmt}(\mathcal{D}^{\mathcal{T}};\theta_\mathcal{M}) = - \sum_{\bm{x},y \in \mathcal{D}}
  y\log p_{\theta_\mathcal{M}}(\hat{y}|\bm{x}_{\text{p}}),
\end{equation}
where $y$ denotes the gold label, and $\theta_\mathcal{M}$ denotes the overall trainable parameters of the PLM.

\begin{figure*}[ht]
  \centering
  \begin{subfigure}[b]{0.48\linewidth}
    \centering
    \includegraphics[page=5,width=\linewidth]{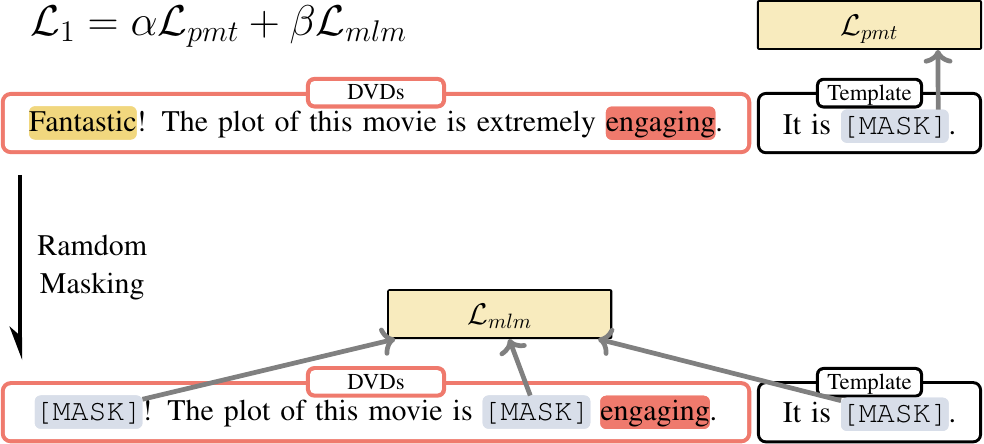}
    \caption{Stage \ref{alg1}: Learn from the source domain.}
  \end{subfigure}\hfill
  \begin{subfigure}[b]{0.48\linewidth}
    \centering
    \includegraphics[page=6,width=\linewidth]{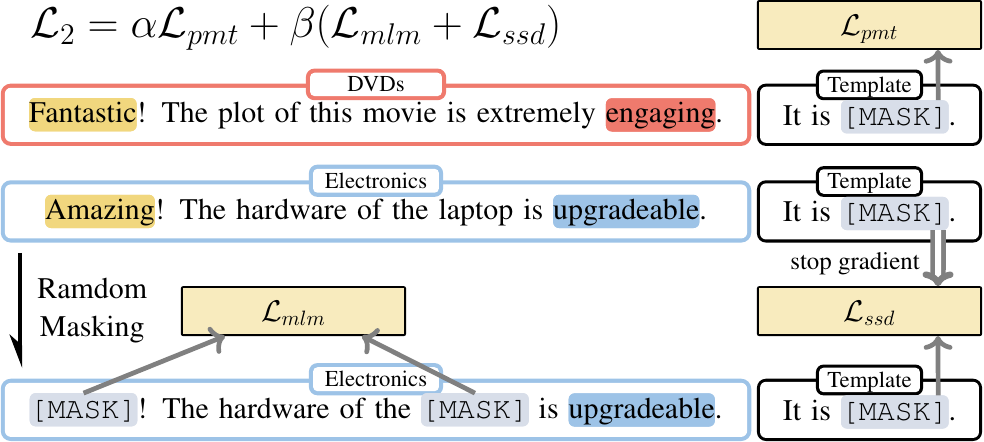}
    \caption{Stage \ref{alg2}: Adapt to the target domain.}
  \end{subfigure}
  \caption{
    An overview of the proposed method.
    The \textcolor{gray}{$\rightarrow$} means the model's output.
    The \textcolor{gray}{$\Rightarrow$} means the model's output which has no gradient.
    The $\mathcal{L}_{pmt}, \mathcal{L}_{mlm}, \mathcal{L}_{ssd}$ mean the losses of prompt tuning for classification, mask language modeling, and self-supervised distillation.
    The training process of the method consists of two stages.
    (a) In Stage \ref{alg1}, we apply mask language modeling and classification task on the source domain, to prevent the model from over-focusing on overfitting features or shortcut features.
    (b) In Stage \ref{alg2}, we continually train the classification task in the source domain, while we do mask language modeling and self-supervised distillation in the target domain.
  } \label{method}
\end{figure*}

\subsection{Mask Language Modeling}

We use mask language modeling to avoid shortcut learning\footnote{Shortcuts learning means that model learned decision rules~(or shortcut features) that perform well on standard benchmarks but fail to transfer to more challenging testing conditions, such as real-world scenarios.} \cite{2020arXiv200407780G} and adapt to the distribution of the target domain.
Specifically, we construct a \textit{masked} textual prompt $\bm{x}_{\text{pm}}$ which is similar to the textual prompt $\bm{x}_{\text{p}}$ in Eq \ref{textual-prompt}.
The masked textual prompt consists of a \textit{masked} input sentence, a template containing \texttt{[MASK]}, and two special tokens (\texttt{[CLS]} and \texttt{[SEP]}):
\begin{equation}\label{masked-textual-prompt}
  \bm{x}_{\text{pm}} = ``\texttt{[CLS]}~\bm{x}_{\text{m}}\text{. It is }\texttt{[MASK]}\text{. }\texttt{[SEP]}",
\end{equation}
where $\bm{x}_{\text{m}}$ is the masked version of $\bm{x}$ in Eq \ref{textual-prompt}.

Here we fine-tune the PLM through the masked language modeling task on $\bm{x}_{\text{m}}$. Given an labeled dataset $\mathcal{D}^{\mathcal{T}} = \{x_i,y_i\}^n_{i=1}$, the loss of each sentence in $\mathcal{D}^{\mathcal{T}}$ is the mean masked LM likelihood of all \texttt{[MASK]} in the sentence.
Furthermore, the overall loss of $\mathcal{D}^{\mathcal{T}}$ is the summation of the individual sentence loss in the dataset:
\begin{gather}
  \scalebox{1.0}{
  $\mathcal{L}_{mlm}(\mathcal{D};\theta_\mathcal{M}) = - \sum_{\bm{x} \in \mathcal{D}} \sum_{\hat{x} \in m(\bm{x}_{\text{m}})} \frac{\log p_{\theta_\mathcal{M}}(\hat{x}|\bm{x}_{\text{pm}})}{len_{m(\bm{x}_{\text{m}})}},$
  }
\end{gather}
where $m(y_{\text{m}})$ and $len_{m(\bm{x}_{\text{m}})}$ denote the masked words and counts in $\bm{x}_{\text{m}}$, respectively, $\theta_\mathcal{M}$ denotes the overall trainable parameters of the PLM.

\section{Method}\label{sec:method}

In this section, we introduce the self-supervised distillation method and the two-stage learning procedure for cross-domain text classification.
Figure \ref{method} illustrates the overall learning procedure.

\subsection{\textbf{S}elf-\textbf{S}upervised \textbf{D}istillation (SSD)}

The method needs a trained model which can perform the task.
In addition, the model has the ability to generalize to the target domain through domain-invariant features, but not all useful features for the task are domain-invariant.
We use the model itself to construct a soft self-supervised signal.
This signal enables the model to establish a connection between predictions and latent domain-aware features of the target domain.
This process, which we refer to as \textbf{S}elf-\textbf{S}upervised \textbf{D}istillation (SSD), constitutes one of our core contributions.

During prediction, the model can only utilize the unmasked features of a masked sentence ($\bm{x}_{\text{pm}}$), but it can use all features of the original version of the sentence ($\bm{x}_{\text{p}}$).
The model will be forced to make the connection between predictions of $\bm{x}_{\text{p}}$ in Eq \ref{textual-prompt} and the unmasked words of $\bm{x}_{\text{pm}}$, which can include domain-invariant, domain-aware features, or both in the target domain.
Recall that they contain the original ($\bm{x}_{\text{p}}$) and masked versions ($\bm{x}_{\text{pm}}$) of the same sentence ($\bm{x}$), respectively.
We perform knowledge distillation between the model predictions of $p_{\theta}(y|\bm{x}_{\text{pm}})$ and $p_{\theta}(y|\bm{x}_{\text{p}})$.
The objective of SSD can be defined as the following formula:
\begin{gather}
  \scalebox{0.9}{
  $\mathcal{L}_{ssd}(\mathcal{D};\theta_\mathcal{M}) = \sum_{\bm{x} \in \mathcal{D}}\mathrm{KL}( p_{\theta_\mathcal{M}}(y|\bm{x}_{\text{pm}}) || p_{\theta_\mathcal{M}}(y|\bm{x}_{\text{p}})),$
  }
\end{gather}
where $\bm{x}_{\text{m}}$ and $\bm{x}_{\text{pm}}$ is processed from the same input sentence $\bm{x}$.

\subsection{Learning Procedure}

Our learning procedure comprises two stages, which are summarized in Algorithm \ref{alg1} and Algorithm \ref{alg2}.
We use the vanilla prompt tuning method without masking during inference.

\subsubsection{Stage \ref{alg1}: Learn from the source domain}\label{sec:tune}

\begin{algorithm}[t]
  \caption{\begin{small}Stage \ref{alg1}: Learn from the source domain\end{small}}
  \label{alg1}
  \begin{algorithmic}[1]
    \INPUT Training samples of source domain labeled dataset $\mathcal{D}_S^{\mathcal{T}}$
    \OUTPUT Configurations of finetuned model $\theta_\mathcal{M}$
    \INITIALIZE PLM $\theta_\mathcal{M}$; learning rate $\eta$; trade-off parameter $\alpha,\beta$

    \While{Training epoch not end}
    \For{$x$ in $\mathcal{D}_S^{\mathcal{T}}$}
    \LineComment{Minimizing the classification loss in source domain}
    \State $\mathcal{L}_1' \gets \alpha \mathcal{L}_{pmt}(\bm{x};\theta_\mathcal{M})$
    \State $\theta_\mathcal{M} = \theta_\mathcal{M} - \eta \nabla_{\theta_\mathcal{M}}\mathcal{L}'$
    \LineComment{Minimizing the MLM loss in source domain}
    \State $\mathcal{L}_1'' \gets \beta \mathcal{L}_{mlm}(\bm{x};\theta_\mathcal{M})$
    \State $\theta_\mathcal{M} = \theta_\mathcal{M} - \eta \nabla_{\theta_\mathcal{M}}\mathcal{L}''$
    \EndFor
    \EndWhile
  \end{algorithmic}
\end{algorithm}

In Stage \ref{alg1}, our objective is to obtain a fine-tuned model with the ability to perform the downstream task effectively.
We use a variant of the mask language modeling as an auxiliary task for prompt tuning to prevent the model from over-focusing overfitting features or shortcut features \cite{2020arXiv200407780G}.
We refer to this approach as \textbf{M}\begin{small}LM\end{small} \textbf{E}nhanced \textbf{P}rompt \textbf{T}uning (MEPT).

We initialize the parameters of the model $\mathcal{M}$ using a Pretrained Language Model (PLM) such as BERT or RoBERTa.
During each iteration, we sample new examples from the source domain $\mathcal{D}_S^{\mathcal{T}}$ to train the model $\mathcal{M}$.
Firstly, we calculate the classification loss of those sentences and update the parameters with the loss, as shown in line 5 of Algorithm \ref{alg1}.
Then we mask the same sentence and calculate mask language modeling loss to update the parameters, as depicted in line 8 of Algorithm \ref{alg1}.
The parameters of the model will be updated together by these two losses.

In summary, the objective of Stage \ref{alg1}, given a labeled dataset $\mathcal{D}^{\mathcal{T}}$, is obtained using the weighted cross-entropy loss for prompt tuning classification ($\mathcal{L}_{pmt}$) and mask language modeling loss ($\mathcal{L}_{mlm}$):
\begin{equation}\label{obj-stage1}
  \begin{aligned}
    \mathcal{L}_1'(\mathcal{D}^{\mathcal{T}};\theta_\mathcal{M})  & = \alpha \mathcal{L}_{pmt}(\mathcal{D}^{\mathcal{T}};\theta_\mathcal{M}), \\
    \mathcal{L}_1''(\mathcal{D}^{\mathcal{T}};\theta_\mathcal{M}) & = \beta \mathcal{L}_{mlm}(\mathcal{D};\theta_\mathcal{M}),
  \end{aligned}
\end{equation}
where $\alpha,\beta$ is the loss weight. As the Algorithm \ref{alg1} shows, we alternate between $\mathcal{L}_{1}'$ and $\mathcal{L}_{1}''$ optimizations during training.

\subsubsection{Stage \ref{alg2}: Adapt to the target domain}\label{sec:adapt}

In Stage \ref{alg2}, we adapt the model trained in Stage \ref{alg1} to the target domain. We refer to the resulting model \textbf{T}wo-stage \textbf{A}dapted \textbf{M}\begin{small}LM\end{small} \textbf{E}nhanced \textbf{P}rompt \textbf{T}uning (\textsc{TamePT}), which is our proposed model.

We initialize the parameters of our \textsc{TamePT} model using the model already tuned in Stage \ref{alg1}.
Firstly, we sample labeled data from the source domain $\mathcal{D}_S^{\mathcal{T}}$ and calculate sentiment classification loss.
The model parameters are updated using this loss in line 5 of Algorithm \ref{alg2}.
Next, we sample unlabeled data from the target domain $\mathcal{D}_T$ and mask the unlabeled data to do a masking language model and self-supervised distillation with the previous prediction.
It should be noted that the self-supervised distillation requires obtaining the prediction of the original sentence before masking.
Finally, the model parameters are updated using the mask language modeling loss and self-supervised distillation loss of target domain examples, as shown in line 8 of Algorithm \ref{alg2}.
The model parameters are updated together using the three aforementioned losses.

\begin{algorithm}[t]
  \caption{\begin{small}Stage  \ref{alg2}: Adapt to the target domain\end{small}}
  \label{alg2}
  \begin{algorithmic}[1]
    \INPUT Training samples of source domain labeled dataset $\mathcal{D}_S^{\mathcal{T}}$ and target domain dataset $\mathcal{D}_T$
    \OUTPUT Configurations of Final Model $\theta_\mathcal{M}$
    \INITIALIZE Model $\theta_\mathcal{M}$ already tuned in Stage \ref{alg1}; learning rate $\eta$; trade-off parameter $\alpha,\beta$

    \While{Training epoch not end}
    \For{$x^s,x^t$ in $\mathcal{D}_S^{\mathcal{T}},\mathcal{D}_T$}
    \LineComment{Minimizing the classification loss in source domain}
    \State $\mathcal{L}_2' \gets \alpha \mathcal{L}_{pmt}(\bm{x}^s;\theta_\mathcal{M})$
    \State $\theta_\mathcal{M} = \theta_\mathcal{M} - \eta \nabla_{\theta_\mathcal{M}}\mathcal{L}'$
    \LineComment{Minimizing the SSD loss and MLM in target domain}
    \State {\small $\mathcal{L}_2'' \gets \beta (\mathcal{L}_{mlm}(\bm{x}^t;\theta_\mathcal{M}) + \mathcal{L}_{ssd}(\bm{x}^t;\theta_\mathcal{M}))$}
    \State $\theta_\mathcal{M} = \theta_\mathcal{M} - \eta \nabla_{\theta_\mathcal{M}}\mathcal{L}''$
    \EndFor
    \EndWhile
  \end{algorithmic}
\end{algorithm}

In conclusion, the training objective for Stage \ref{alg2} is obtained using the weighted cross-entropy loss for classification ($\mathcal{L}_{pmt}$), mask language modeling loss ($\mathcal{L}_{mlm}$) and self-supervised distillation loss ($\mathcal{L}_{ssd}$). Given an labeled dataset $\mathcal{D}^{\mathcal{T}}_S$ and unlabeled datasets $\mathcal{D}_T$.
% \chadded{
the loss can be defined as:
\begin{equation}
  \begin{aligned}
    \mathcal{L}_2'(\mathcal{D}^{\mathcal{T}}_S,\mathcal{D}_T;\theta_\mathcal{M})  & = \alpha \mathcal{L}_{pmt}(\mathcal{D}^{\mathcal{T}}_S;\theta_\mathcal{M}), \\
    \mathcal{L}_2''(\mathcal{D}^{\mathcal{T}}_S,\mathcal{D}_T;\theta_\mathcal{M}) & =  \beta (\mathcal{L}_{mlm}(\mathcal{D}_T;\theta_\mathcal{M})               \\
                                                                                  & + \mathcal{L}_{ssd}(\mathcal{D}_T;\theta_\mathcal{M})),
  \end{aligned}
  \label{obj-stage2}
\end{equation}
where $\alpha,\beta$ is the loss weight.
As the Algorithm \ref{alg2} shows, we alternate between $\mathcal{L}_{2}'$ and $\mathcal{L}_{2}''$ optimizations during training.

\subsection{Summary}
The proposed method consists of two stages, as shown in Figure \ref{method}.
\textbf{M}\begin{small}LM\end{small} \textbf{E}nhanced \textbf{P}rompt \textbf{T}uning (MEPT) means that we use mask language modeling as an auxiliary task for prompt tuning.
\textbf{T}wo-stage \textbf{A}dapted \textbf{M}\begin{small}LM\end{small} \textbf{E}nhanced \textbf{P}rompt \textbf{T}uning (\textsc{TamePT}) means that we use the model tuned in the source domain to adapt to the target domain with mask language modeling and self-supervised distillation.

% \begin{table}
%   \begin{tabular}[t]{ll}
%     MEPT & \textbf{M}ask \textbf{L}anguage \textbf{M}odeling \textbf{E}nhanced \textbf{P}rompt \textbf{T}uning \\
%     \textsc{TamePT} & \textbf{T}wo-stage \textbf{A}dapted \textbf{M}ask \textbf{L}anguage \textbf{M}odeling \textbf{E}nhanced \textbf{P}rompt \textbf{T}uning \\
%   \end{tabular}
% \end{table}

\section{Experiments}

In this section, we begin by introducing the standard benchmark used for cross-domain text classification on which we conduct experiments.
We then present a series of baseline models that we use for comparison purposes.
Subsequently, we provide a detailed description of our method's training procedure.
Finally, we present the results of the experiment and the analysis of our method.

\subsection{Dataset}
We evaluate the effectiveness of our proposed  method on the Amazon reviews dataset \citetlanguageresource{blitzer-etal-2007-biographies}, which is a widely-used benchmark dataset for cross-domain text classification.
The dataset contains reviews in four different domains: Books (B), DVDs (D), Electronics (E), and Kitchen appliances (K) with 2,000 manually labeled reviews in each domain, equally balanced between positive and negative sentiments.
Additionally, the dataset also provides a certain amount of unlabeled data for each domain, as shown in Table \ref{tab:amazon}.

\textbf{(1)} In the single-source domain adaptation experiments, we adopt the setting used in \cite{karouzos-etal-2021-udalm} and \cite{wu-shi-2022-adversarial} to construct 12 cross-domain text classification tasks, each corresponding to a distinct ordered domain pair.
In each of these 12 adaptation scenarios, we apply 20\% of both labeled source and unlabeled target data for validation,
while the labeled target data are used exclusively for testing and are not seen during training or validation.
\textbf{(2)} In the multi-source domain adaptation experiments, we follow \cite{wu-shi-2022-adversarial} to construct 4 cross-domain text classification tasks.
Specifically, we choose one as the target domain and the remaining three domains as multiple source domains, resulting in tasks such as ``BDE $\rightarrow$ K'', ``BDK $\rightarrow$ E''.

\begin{table}[t]
  \centering
  \resizebox{\linewidth}{!}{%
    \begin{tabular}{crrr}
      \toprule
      \textbf{Domain}          & \textbf{\# Positive} & \textbf{\# Negative} & \textbf{\# Unlabeled} \\ \midrule
      \textbf{B}ooks (B)       & 1,000                & 1,000                & 6,000                 \\
      \textbf{D}VDs (D)        & 1,000                & 1,000                & 34,741                \\
      \textbf{E}lectronics (E) & 1,000                & 1,000                & 13,153                \\
      \textbf{K}itchen (K)     & 1,000                & 1,000                & 16,785                \\ \bottomrule
    \end{tabular}%
  }
  \caption{Statistics for the Amazon reviews multi-domain classification dataset.}
  \label{tab:amazon}
\end{table}

\subsection{Baselines}
We present several strong baselines in our experiments and demonstrate the effectiveness of our proposed methods.

\begin{enumerate}
  \item \textbf{R-PERL} \cite{10.1162/tacl_a_00328}: Utilize BERT for cross-domain text classification with pivot-based fine-tuning.
  \item \textbf{DAAT} \cite{du-etal-2020-adversarial}: Employ BERT post-training for cross-domain text classification through adversarial training.
  \item \textbf{p+CFd} \cite{ye-etal-2020-feature}: Leverage XLM-R for cross-domain text classification employing class-aware feature self-distillation~(CFd).
  \item \textbf{SENTIX$_{Fix}$} \cite{zhou-etal-2020-sentix}: Pre-train a sentiment-aware language model via multiple pre-training tasks.
  \item \textbf{UDALM} \cite{karouzos-etal-2021-udalm}: Conduct fine-tuning with a mixed classification and MLM loss on domain-adapted PLMs.
  \item \textbf{AdSPT} \cite{wu-shi-2022-adversarial}: Execute soft prompt tuning with an adversarial training object on vanilla PLMs.
\end{enumerate}

We present our proposed method, denoted as \textbf{\textsc{TamePT}}, which is comprehensively introduced in Section \ref{sec:method}.
To thoroughly evaluate the performance, in alignment with \cite{karouzos-etal-2021-udalm} and \cite{wu-shi-2022-adversarial}, we employ \textit{accuracy} as the selected evaluation metric.

\subsection{Implementation Details}
We adopt a 12-layer Transformer \cite{NIPS2017_3f5ee243,devlin-etal-2019-bert} initialized with RoBERTa$_{base}$ \cite{DBLP:journals/corr/abs-1907-11692} as the PLM.

\textbf{(1)} During Stage \ref{alg1}, we conduct training over 10 epochs with a batch size of 4, employing early stopping (patience = 3) based on the accuracy metric. The chosen optimizer for this phase is AdamW \cite{DBLP:journals/corr/abs-1711-05101}, with a learning rate of $1 \times 10^{-5}$.
Additionally, we implement a strategy to halve the learning rate every 3 epochs.
For this stage, we set $\alpha=1.0,\beta=0.6$ for Eq. \ref{obj-stage1}.
\textbf{(2)} During Stage \ref{alg2}, we conduct training over 10 epochs with a batch size of 4, with early stopping (patience = 3) on the mixing loss encompassing both classification loss and mask language modeling loss.
The optimization utilizes AdamW with a learning rate of $1 \times 10^{-6}$ without learning rate decay.
The parameter settings for this stage are adjusted to $\alpha=0.5,\beta=0.5$ for Eq. \ref{obj-stage2}.

Furthermore, for both the mask language modeling objective and the self-supervised distillation objective, we adopt a strategy wherein 30\% of tokens are randomly replaced with \texttt{[MASK]} or random tokens. To manage the input size, the maximum sequence length is set to 512 through truncation of inputs.
In a notable measure, during Stage \ref{alg2}, we randomly select an equal number of unlabeled data from the target domain of every epoch.

\begin{table*}[htbp]
  \centering
  \begin{adjustbox}{width=0.95\linewidth}
    \begin{tabular}{cccccccc}
      \toprule
      S → T & R-PERL & DAAT  & p+CFd                  & UDALM                  & SENTIX$_{Fix}$ & AdSPT          & \textsc{TamePT}                 \\ \midrule
      B → D & 87.80  & 89.70 & 87.65\small{$\pm$0.10} & 90.97\small{$\pm$0.22} & 91.30          & 92.00          & \textbf{93.27}\small{$\pm$0.49} \\
      B → E & 87.20  & 89.57 & 91.30\small{$\pm$0.20} & 91.69\small{$\pm$0.31} & 93.25          & 93.75          & \textbf{94.82}\small{$\pm$0.23} \\
      B → K & 90.20  & 90.75 & 92.45\small{$\pm$0.60} & 93.21\small{$\pm$0.22} & \textbf{96.20} & 93.10          & 95.75\small{$\pm$0.40}          \\
      D → B & 85.60  & 90.86 & 91.50\small{$\pm$0.40} & 91.00\small{$\pm$0.42} & 91.15          & 92.15          & \textbf{94.83}\small{$\pm$0.31} \\
      D → E & 89.30  & 89.30 & 91.55\small{$\pm$0.30} & 92.30\small{$\pm$0.47} & 93.55          & 94.00          & \textbf{94.57}\small{$\pm$0.18} \\
      D → K & 90.40  & 87.53 & 92.45\small{$\pm$0.20} & 93.66\small{$\pm$0.37} & \textbf{96.00} & 93.25          & 95.84\small{$\pm$0.24}          \\
      E → B & 90.20  & 88.91 & 88.65\small{$\pm$0.40} & 90.61\small{$\pm$0.30} & 90.40          & 92.70          & \textbf{93.20}\small{$\pm$0.63} \\
      E → D & 84.80  & 90.13 & 88.20\small{$\pm$0.40} & 88.83\small{$\pm$0.61} & 91.20          & \textbf{93.15} & 92.63\small{$\pm$0.34}          \\
      E → K & 91.20  & 93.18 & 93.60\small{$\pm$0.50} & 94.43\small{$\pm$0.24} & \textbf{96.20} & 94.75          & 96.16\small{$\pm$0.07}          \\
      K → B & 83.00  & 87.98 & 89.75\small{$\pm$0.80} & 90.29\small{$\pm$0.51} & 89.55          & \textbf{92.35} & 92.18\small{$\pm$0.84}          \\
      K → D & 85.60  & 88.81 & 87.80\small{$\pm$0.40} & 89.54\small{$\pm$0.59} & 89.85          & \textbf{92.55} & 91.77\small{$\pm$0.68}          \\
      K → E & 91.20  & 91.72 & 92.60\small{$\pm$0.50} & 94.34\small{$\pm$0.26} & 93.55          & 93.95          & \textbf{95.06}\small{$\pm$0.43} \\ \midrule
      AVG   & 87.50  & 90.12 & 90.63\small{$\pm$0.40} & 91.74\small{$\pm$0.38} & 92.68          & 93.14          & \textbf{94.17}\small{$\pm$0.40} \\ \bottomrule
    \end{tabular}%
  \end{adjustbox}
  \caption{
    Results of single-source domain adaptation on Amazon reviews.
    There are four domains, B: Books, D: DVDs, E: Electronics, K: Kitchen appliances.
    In the table header, S: Source domain; T: Target domain.
    The \textsc{TamePT} is our proposed method, which is described in Section \ref{sec:method}.
    We report mean performances and standard errors over 5 seeds.
  }
  \label{main-experiment}
\end{table*}

\begin{table}[ht]
  \centering
  \begin{adjustbox}{width=0.8\linewidth}
    \begin{tabular}{ccc}
      \toprule
      S → T   & AdSPT & \textsc{TamePT}                 \\ \midrule
      BDE → K & 93.75 & \textbf{96.13}\small{$\pm$0.12} \\
      BDK → E & 94.25 & \textbf{95.68}\small{$\pm$0.11} \\
      BEK → D & 93.50 & \textbf{93.98}\small{$\pm$0.16} \\
      DEK → B & 93.50 & \textbf{94.57}\small{$\pm$0.37} \\ \midrule
      AVG     & 93.75 & \textbf{95.09}\small{$\pm$0.19} \\ \bottomrule
    \end{tabular}%
  \end{adjustbox}
  \caption{
    Results of multi-source domain adaptation on Amazon reviews.
    The AdSPT is the only one of the baselines that do experiments in the multi-source domain adaptation settings.
    We report mean performances and standard errors over 5 seeds.
  }
  \label{multi-domain-main-experiment}
\end{table}

All the models and the accompanying analysis are meticulously implemented utilizing the PyTorch framework \cite{NEURIPS2019_9015}, along with Hydra framework \cite{Yadan2019Hydra}, PyTorch Lightning \cite{Falcon_PyTorch_Lightning_2019}, HuggingFace transformers \cite{wolf-etal-2020-transformers} and datasets \cite{lhoest-etal-2021-datasets} for a streamlined workflow.

\subsection{Experiment Results}

In this section, we concentrate on the results of single-source domain adaptation (Table \ref{main-experiment}) and multi-source domain adaptation (Table \ref{multi-domain-main-experiment}).
Our proposed method, \textsc{TamePT}, achieves new state-of-the-art performance for both tasks, with an accuracy of 94.17\% (+1.03\%) for single-source domain adaptation and 95.09\% (+1.34\%) for multi-source domain adaptation. These results demonstrate the effectiveness of our approach for adapting PLMs to cross-domain text classification tasks.

\subsubsection{Single-Source Domain Adaptation}

The main experiment results in Table \ref{main-experiment} demonstrate that our proposed method, \textsc{TamePT}, outperforms other state-of-the-art methods in most single-source domain adaptation settings. Specifically, compared to previous state-of-the-art methods, \textsc{TamePT} achieves significantly higher average accuracy (1.03\% absolute improvement over AdSPT, 1.49\% absolute improvement over SENTIX${_{Fix}}$, 2.43\% absolute improvement over UDALM, and 4.05\% absolute improvement over DAAT).
However, the AdSPT achieves better performance in experiments ``E → D'' and ``K → D''.
Furthermore, SENTIX${_{Fix}}$ achieves the best performance when the target domain is ``K'', but our method still achieves comparable performance.
It is mainly because the extra training data of SENTIX${_{Fix}}$ is closer to the domain ``K''.

\begin{table*}[htbp]
  \centering
  \begin{adjustbox}{width=0.8\textwidth}
    \begin{tabular}{cc|cc|cc}
      \toprule
      ~     & ~                               & \multicolumn{2}{c|}{Stage \ref{alg1}} & \multicolumn{2}{c}{Stage \ref{alg2}}                                                                     \\
      S → T & \textsc{TamePT}                 & w/o Stage \ref{alg2}                  & w/o mlm                              & w/o mlm                         & w/o ssd                         \\ \midrule
      B → D & \textbf{93.27}\small{$\pm$0.49} & 92.88\small{$\pm$0.37}                & 92.77\small{$\pm$0.19}               & 93.21\small{$\pm$0.36}          & 93.14\small{$\pm$0.11}          \\
      B → E & \textbf{94.82}\small{$\pm$0.23} & 94.30\small{$\pm$0.18}                & 94.22\small{$\pm$0.44}               & 94.60\small{$\pm$0.19}          & 94.29\small{$\pm$0.17}          \\
      B → K & \textbf{95.75}\small{$\pm$0.40} & 95.19\small{$\pm$0.48}                & 94.86\small{$\pm$0.51}               & 95.70\small{$\pm$0.38}          & 95.39\small{$\pm$0.24}          \\
      D → B & \textbf{94.83}\small{$\pm$0.31} & 94.37\small{$\pm$0.29}                & 93.76\small{$\pm$0.42}               & 94.38\small{$\pm$0.36}          & 94.25\small{$\pm$0.31}          \\
      D → E & \textbf{94.57}\small{$\pm$0.18} & 94.11\small{$\pm$0.34}                & 93.82\small{$\pm$0.38}               & 94.47\small{$\pm$0.51}          & 94.33\small{$\pm$0.24}          \\
      D → K & \textbf{95.84}\small{$\pm$0.24} & 94.99\small{$\pm$0.25}                & 94.87\small{$\pm$0.15}               & 95.77\small{$\pm$0.17}          & 95.10\small{$\pm$0.31}          \\
      E → B & 93.20\small{$\pm$0.63}          & 92.19\small{$\pm$0.66}                & 92.54\small{$\pm$0.37}               & \textbf{93.24}\small{$\pm$0.23} & 92.70\small{$\pm$0.45}          \\
      E → D & \textbf{92.63}\small{$\pm$0.34} & 90.71\small{$\pm$0.69}                & 91.29\small{$\pm$0.46}               & 92.20\small{$\pm$0.43}          & 91.19\small{$\pm$0.25}          \\
      E → K & \textbf{96.16}\small{$\pm$0.07} & 95.70\small{$\pm$0.58}                & 95.40\small{$\pm$0.46}               & 95.84\small{$\pm$0.19}          & 95.77\small{$\pm$0.23}          \\
      K → B & 92.18\small{$\pm$0.84}          & 92.09\small{$\pm$0.53}                & 91.89\small{$\pm$0.37}               & 91.94\small{$\pm$1.99}          & \textbf{92.55}\small{$\pm$0.26} \\
      K → D & \textbf{91.77}\small{$\pm$0.68} & 90.49\small{$\pm$0.70}                & 89.85\small{$\pm$0.44}               & 89.91\small{$\pm$4.18}          & 91.38\small{$\pm$0.36}          \\
      K → E & 95.06\small{$\pm$0.43}          & 94.55\small{$\pm$0.26}                & 94.58\small{$\pm$0.32}               & \textbf{95.15}\small{$\pm$0.15} & 95.02\small{$\pm$0.04}          \\ \midrule
      AVG   & \textbf{94.17}\small{$\pm$0.40} & 93.46\small{$\pm$0.44}                & 93.32\small{$\pm$0.38}               & 93.87\small{$\pm$0.76}          & 93.76\small{$\pm$0.25}          \\ \bottomrule
    \end{tabular}
  \end{adjustbox}
  \caption{
    The ablation experiments of our method.
    There are four domains, B: Books, D: DVDs, E: Electronics, K: Kitchen appliances.
    In the table header, S: Source domain; T: Target domain.
    The ``mlm'' and ``ssd'' mean self-supervised distillation and mask language modeling.
    The ``w/o Stage \ref{alg2}'' is also called MEPT, which is described in Section \ref{sec:tune}.
    We report mean performances and standard errors over 5 seeds.
  }
  \label{ablation-experiments}
\end{table*}

\subsubsection{Multi-Source Domain Adaptation}

The results presented in Table \ref{multi-domain-main-experiment} demonstrate the superior performance of our proposed \textsc{TamePT} method in all multi-source domain adaptations.
Compared to the previous state-of-the-art model AdSPT, \textsc{TamePT} achieves significantly higher average accuracy (1.34\% absolute improvement).
Notably, our method also achieves better performance than the single-domain adaptation method in most cases, with a lower standard error, indicating its ability to maintain stability as the amount of data and domain increases.
However, when the target domain is ``K'', the result of ``E → K'' (in Table~\ref{main-experiment}) is superior to that of ``BDE → K'' (96.16\% v.s. 96.13\%).
A similar situation occurs in AdSPT (94.75\% v.s. 93.75\%).
It is mainly because the feature distribution of ``E'' and ``K'' is closer.

\subsection{Analysis}

\begin{figure*}[t]
  \centering
  \begin{subfigure}[b]{0.45\linewidth}
    \centering
    \includegraphics[page=8,width=\textwidth]{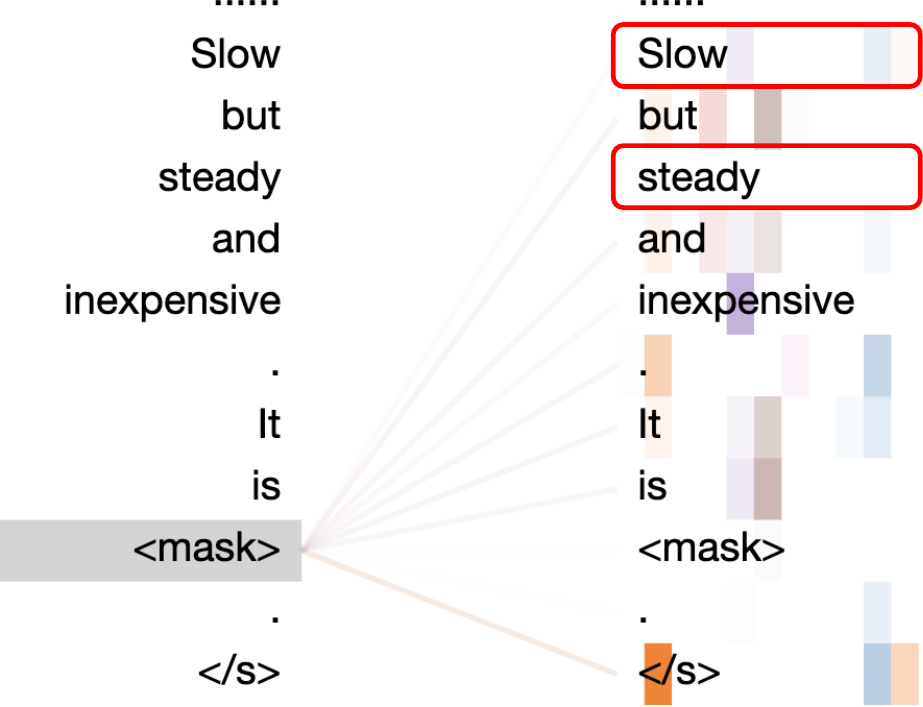}
    \caption{The gradient from the MEPT.}
  \end{subfigure}
  \hfill
  \begin{subfigure}[b]{0.45\linewidth}
    \centering
    \includegraphics[page=9,width=\textwidth]{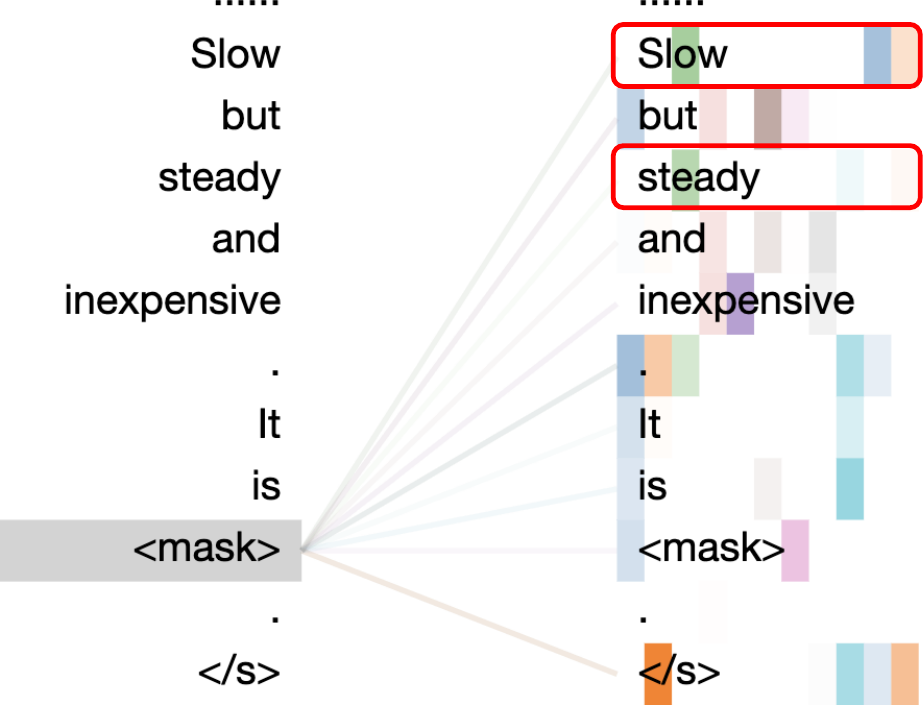}
    \caption{The gradient from the T\begin{tiny}AME\end{tiny}PT.}
  \end{subfigure}
  \caption{
    Visualization for the sentence ``\textit{Plain Vanilla Wireless adapter for you. Slow but steady and inexpensive.}'' in the ``D → E'' setting.
    The different colors mean the gradient of different heads.
    Compared with the gradient from MEPT, the gradient from \textsc{TamePT} pays more attention to the domain-aware features (``slow'' and ``steady'').
  }
  \label{fig:case-study}
\end{figure*}

Table \ref{ablation-experiments} indicates the results of our ablation experiments, which are conducted to evaluate the contributions of different components in our proposed method.
Figure \ref{fig:case-study} shows the case study of our method.
Additionally, Table \ref{generality-experiments} presents the results of experiments that validate the generality of our two-stage adaptation approach on different methods and pre-trained models.
These experiments demonstrate the effectiveness and flexibility of our proposed method.

\subsubsection{Ablation Study}

Our proposed method comprises two stages, namely Stage \ref{alg1} and Stage \ref{alg2}. To evaluate the effectiveness of each stage, we conduct ablation experiments by removing the corresponding components and comparing the performance of the resulting model with the original model. The results of the ablation experiments are presented in Table \ref{ablation-experiments}.

In Stage \ref{alg1}, we observe that the use of \textbf{mask language modeling} is crucial for achieving high accuracy. Specifically, when we remove the mask language modeling, the performance of the Stage \ref{alg1} model dropped by an average of 0.14\% (from 93.46\% to 93.32\%), as shown in Table \ref{ablation-experiments}. With the exception of the ``E → D" experiment, the model including mask language modeling consistently outperforms the one without mask language modeling.

In Stage \ref{alg2}, we find that using both \textbf{self-supervised distillation} and \textbf{mask language modeling} is critical for achieving high accuracy and stability. When we remove the self-supervised distillation, the performance of the Stage \ref{alg2} model decreases by an average of 0.41\% (from 94.17\% to 93.76\%). Similarly, when we remove the mask language modeling, the performance of the Stage \ref{alg2} model decreases by an average of 0.30\% (from 94.17\% to 93.87\%), while the standard error increases by 0.36 (from 0.40 to 0.76). Notably, the standard error for experiments ``K → B" and ``K → D" increases significantly by 1.15 (from 0.84 to 1.99) and 3.50 (from 0.68 to 4.18), respectively. These results indicate that mask language modeling is a crucial factor for achieving high accuracy and, in particular, stability in the proposed method.

In summary, our experiments confirm the effectiveness of both self-supervised distillation and mask language modeling in achieving high accuracy and stability.

\subsubsection{Case Study}

As shown in Figure \ref{fig:case-study}, we conduct a case study to demonstrate the effectiveness of our method in capturing domain-aware features from unlabeled data in the target domain. Specifically, we used the sentence ``\textit{Plain Vanilla Wireless adapter for you. Slow but steady and inexpensive.}'' as an example to analyze the gradients from MEPT\footnote{MEPT: The proposed model T\begin{tiny}AME\end{tiny}PT without Stage \ref{alg2}.} and \textsc{TamePT} in the ``D → E'' setting. The gradients from MEPT and \textsc{TamePT} are depicted in Figures \ref{fig:case-study}(a-b), respectively. Notably, compared to the gradient of MEPT, the gradient from \textsc{TamePT} places more emphasis on the words ``slow" and ``steady", which are domain-aware features. This result demonstrates that our method is capable of assisting models in capturing domain-aware features from unlabeled data in the target domain.

\begin{table}[tbp]
  \centering
  \begin{adjustbox}{width=\linewidth}
    \begin{tabular}{r|cc|cc}
      \toprule
      ~      & \multicolumn{2}{c|}{BERT} & \multicolumn{2}{|c}{RoBERTa}                                                            \\
      Method & Stage \ref{alg1}          & +Stage \ref{alg2}            & Stage \ref{alg1}       & +Stage \ref{alg2}               \\ \midrule
      FT     & 90.00\small{$\pm$0.51}    & 91.10\small{$\pm$1.77}       & 93.31\small{$\pm$0.47} & \textbf{94.09}\small{$\pm$0.38} \\
      PT     & 90.78\small{$\pm$0.78}    & 91.62\small{$\pm$1.59}       & 93.63\small{$\pm$0.36} & \textbf{94.38}\small{$\pm$0.36} \\
      MEPT   & 91.27\small{$\pm$0.58}    & 92.54\small{$\pm$0.26}       & 93.74\small{$\pm$0.43} & \textbf{94.40}\small{$\pm$0.35} \\ \midrule
      AVG    & 91.02\small{$\pm$0.62}    & 91.75\small{$\pm$1.21}       & 93.56\small{$\pm$0.42} & \textbf{94.29}\small{$\pm$0.36} \\ \bottomrule
    \end{tabular}
  \end{adjustbox}
  \caption{
    The experiments for validating the generality of our method.
    We report mean performances and standard error of 12 single-source domain adaptations and 4 multi-domain adaptations.
    The MEPT initialized with RoBERTa and adapted by Stage \ref{alg2} is denoted as ``\textsc{TamePT}'', which is our proposed model.
  }
  \label{generality-experiments}
\end{table}

\subsubsection{Generality Study}

To validate the generality of our method, we conduct experiments on different pre-trained models and methods, as summarized in Table \ref{generality-experiments}. Specifically, we apply Algorithm \ref{alg2} to the fine-tuning (FT), prompt tuning (PT), and MEPT methods. The results demonstrate the effectiveness of our method on different pre-trained models and methods, with an average improvement of 0.73\% (from 91.02\% to 91.75\% with BERT and from 93.56\% to 94.29\% with RoBERTa). These findings support the generality of our proposed method.

% 敏感性分析
\subsubsection{Sensitive Analysis}
\begin{figure}
  \centering
  \includegraphics[width=0.8\linewidth]{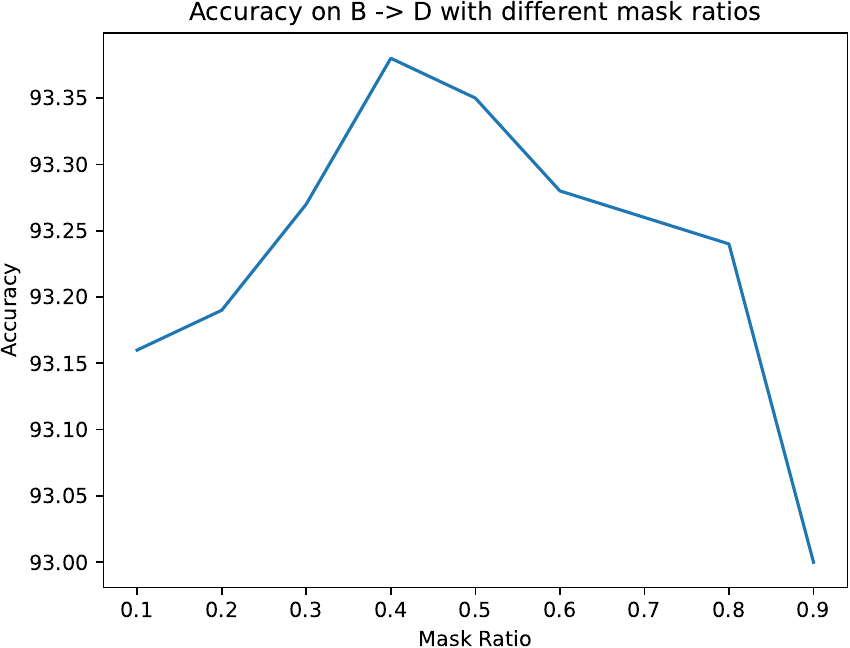}
  \caption{
    The sensitive analysis of the hyperparameters mask ratio in the $B \rightarrow D$ setting.
  }
  \label{fig:sensitive-analysis}
\end{figure}

To evaluate the sensitivity of the hyperparameters, we conduct a sensitive analysis of the mask ratio, which is the ratio of the number of masked tokens to the total number of tokens. The results are shown in Figure \ref{fig:sensitive-analysis}. The results  in the $B \rightarrow D$ setting demonstrate that the performance of our method is relatively stable across a wide range of mask ratios, with the maximum difference in average accuracy being less than 0.5\%. This finding indicates that our method is robust to changes in the mask ratio.

\section{Related Work}

Unsupervised Domain Adaptation is a technique that addresses domain shift issues by learning labeled data of the source domain(s) and unlabeled data of the target domain, which is typically available for both source and the target domains \cite{ramponi-plank-2020-neural}.

As mentioned in Section \ref{sec:intro}, current works can be roughly categorized into two groups.
One group aims to capture domain-invariant features, including pivot-based methods \cite{10.1162/tacl_a_00328}, domain adversarial training \cite{wu-shi-2022-adversarial}, class-aware feature self-distillation \cite{ye-etal-2020-feature}, and sentiment-aware language model \cite{zhou-etal-2020-sentix}.
Another group employs pre-trained models to exploit the task-agnostic features of the target domain during domain adaptation by language modeling \cite{du-etal-2020-adversarial, karouzos-etal-2021-udalm}.
Some works use both domain-invariant and task-agnostic features \cite{du-etal-2020-adversarial}.
However, previous work only focuses on extracting domain-invariant features or task-agnostic features.
In contrast, we are the first to consider the domain-aware features of the target domain.

\section{Conclusion}

In this paper, we propose a two-stage learning procedure for cross-domain text classification that leverages self-supervised distillation to capture domain-aware features in the target domain.
We demonstrate that this procedure outperforms previous state-of-the-art models in most cases, achieving a significant improvement in average accuracy.
Our experiments also highlight the significance of self-supervised distillation and mask language modeling in achieving high performance and stability.
Moreover, the two-stage learning procedure can be easily applied to existing trained models for cross-domain text classification.

\section*{Acknowledgments}
We gratefully acknowledge the support of the National Natural Science Foundation of China (NSFC) via grant 62236004 and 62206078, and the support of Du Xiaoman (Beijing) Science Technology Co., Ltd.

\section{Bibliographical References}\label{sec:reference}

\bibliographystyle{lrec-coling2024-natbib}
\bibliography{anthology,custom}

\section{Language Resource References}\label{lr:ref}
\bibliographystylelanguageresource{lrec-coling2024-natbib}
\bibliographylanguageresource{languageresource}

\appendix

\onecolumn

\section{Details of the Experiments}\label{sec:appendix-experiments}
The details of the experiments in cross-domain text classification on Amazon reviews,
including 6 models, two pre-trained models~(BERT$_{base}$ and RoBERTa$_{base}$),
12 single-source domain adaptations and 4 multi-source domain adaptations.

\begin{table*}[ht]
  \centering
  \begin{adjustbox}{width=\textwidth}
    \begin{tabular}{ccccccc}
      \toprule
      S → T   & FT                     & PT                     & MEPT                   & FT+stage \ref{alg2}     & PT+stage \ref{alg2}             & \ours                           \\ \midrule
      B → D   & 90.57\small{$\pm$0.35} & 90.60\small{$\pm$0.46} & 90.62\small{$\pm$0.34} & 90.20\small{$\pm$0.27}  & 90.72\small{$\pm$0.27}          & \textbf{91.01}\small{$\pm$0.25} \\
      B → E   & 90.71\small{$\pm$0.71} & 90.80\small{$\pm$1.30} & 90.88\small{$\pm$0.80} & 91.87\small{$\pm$0.51}  & \textbf{92.24}\small{$\pm$0.72} & 92.05\small{$\pm$0.16}          \\
      B → K   & 91.86\small{$\pm$0.42} & 92.07\small{$\pm$0.42} & 92.51\small{$\pm$0.28} & 92.81\small{$\pm$0.80}  & 93.53\small{$\pm$0.47}          & \textbf{94.06}\small{$\pm$0.23} \\
      D → B   & 91.10\small{$\pm$0.96} & 90.61\small{$\pm$0.67} & 90.83\small{$\pm$0.58} & 91.71\small{$\pm$0.91}  & 91.14\small{$\pm$0.45}          & \textbf{91.76}\small{$\pm$0.20} \\
      D → E   & 90.20\small{$\pm$0.58} & 90.38\small{$\pm$1.10} & 90.71\small{$\pm$0.80} & 91.65\small{$\pm$0.28}  & 92.21\small{$\pm$0.13}          & \textbf{92.98}\small{$\pm$0.14} \\
      D → K   & 91.43\small{$\pm$0.43} & 91.28\small{$\pm$0.66} & 92.05\small{$\pm$0.43} & 92.89\small{$\pm$0.59}  & 92.95\small{$\pm$0.58}          & \textbf{93.72}\small{$\pm$0.26} \\
      E → B   & 89.42\small{$\pm$0.98} & 87.71\small{$\pm$1.66} & 88.64\small{$\pm$1.62} & 90.05\small{$\pm$0.48}  & 91.22\small{$\pm$0.30}          & \textbf{91.72}\small{$\pm$0.32} \\
      E → D   & 88.77\small{$\pm$0.88} & 88.56\small{$\pm$1.11} & 88.73\small{$\pm$0.46} & 88.99\small{$\pm$0.47}  & \textbf{90.27}\small{$\pm$0.13} & 90.03\small{$\pm$0.31}          \\
      E → K   & 94.34\small{$\pm$0.34} & 92.64\small{$\pm$1.22} & 93.76\small{$\pm$0.94} & 94.47\small{$\pm$0.48}  & 94.60\small{$\pm$0.49}          & \textbf{94.90}\small{$\pm$0.37} \\
      K → B   & 89.02\small{$\pm$0.67} & 88.37\small{$\pm$1.22} & 89.52\small{$\pm$0.47} & 90.05\small{$\pm$0.51}  & 90.76\small{$\pm$0.51}          & \textbf{91.13}\small{$\pm$0.36} \\
      K → D   & 88.13\small{$\pm$0.14} & 88.43\small{$\pm$0.59} & 88.49\small{$\pm$0.65} & 87.72\small{$\pm$1.16}  & 89.40\small{$\pm$0.46}          & \textbf{90.27}\small{$\pm$0.25} \\
      K → E   & 92.53\small{$\pm$0.29} & 92.43\small{$\pm$0.53} & 92.77\small{$\pm$0.36} & 92.34\small{$\pm$1.24}  & 93.13\small{$\pm$0.15}          & \textbf{93.42}\small{$\pm$0.22} \\ \midrule
      AVG     & 90.67\small{$\pm$0.56} & 90.32\small{$\pm$0.91} & 90.79\small{$\pm$0.64} & 91.23\small{$\pm$0.64}  & 91.85\small{$\pm$0.39}          & \textbf{92.25}\small{$\pm$0.26} \\ \midrule
      BDE → K & 93.87\small{$\pm$0.31} & 93.88\small{$\pm$0.24} & 94.48\small{$\pm$0.24} & 95.06\small{$\pm$0.36}  & 95.13\small{$\pm$0.55}          & \textbf{95.36}\small{$\pm$0.30} \\
      BDK → E & 92.78\small{$\pm$0.48} & 93.00\small{$\pm$0.49} & 93.42\small{$\pm$0.36} & 84.80\small{$\pm$19.43} & 84.93\small{$\pm$19.53}         & \textbf{93.98}\small{$\pm$0.16} \\
      BEK → D & 90.49\small{$\pm$0.37} & 90.66\small{$\pm$0.54} & 91.17\small{$\pm$0.43} & 91.00\small{$\pm$0.55}  & 91.52\small{$\pm$0.29}          & \textbf{91.61}\small{$\pm$0.30} \\
      DEK → B & 90.92\small{$\pm$0.23} & 91.06\small{$\pm$0.33} & 91.72\small{$\pm$0.49} & 91.98\small{$\pm$0.25}  & 92.24\small{$\pm$0.41}          & \textbf{92.65}\small{$\pm$0.30} \\ \midrule
      AVG     & 92.01\small{$\pm$0.35} & 92.15\small{$\pm$0.40} & 92.70\small{$\pm$0.38} & 90.71\small{$\pm$5.15}  & 90.95\small{$\pm$5.20}          & \textbf{93.40}\small{$\pm$0.27} \\ \bottomrule
    \end{tabular}%
  \end{adjustbox}
  \caption{
    Results on Amazon reviews based on BERT$_{base}$.
  }
  \label{bert-detailed-experiment}
\end{table*}

\begin{table*}[htbp]
  \centering
  \begin{adjustbox}{width=\textwidth}
    \begin{tabular}{ccccccc}
      \toprule
      S → T   & FT                     & PT                     & MEPT                   & FT+stage \ref{alg2}             & PT+stage \ref{alg2}             & \ours                           \\ \midrule
      B → D   & 92.67\small{$\pm$0.35} & 92.77\small{$\pm$0.19} & 92.88\small{$\pm$0.37} & 92.96\small{$\pm$0.19}          & 93.23\small{$\pm$0.45}          & \textbf{93.27}\small{$\pm$0.49} \\
      B → E   & 93.94\small{$\pm$0.12} & 94.22\small{$\pm$0.44} & 94.30\small{$\pm$0.18} & 94.47\small{$\pm$0.15}          & 94.80\small{$\pm$0.34}          & \textbf{94.82}\small{$\pm$0.23} \\
      B → K   & 94.59\small{$\pm$0.33} & 94.86\small{$\pm$0.51} & 95.19\small{$\pm$0.48} & 95.71\small{$\pm$0.39}          & \textbf{95.78}\small{$\pm$0.49} & 95.75\small{$\pm$0.40}          \\
      D → B   & 93.78\small{$\pm$0.37} & 93.76\small{$\pm$0.42} & 94.37\small{$\pm$0.29} & 94.28\small{$\pm$0.48}          & 94.18\small{$\pm$0.33}          & \textbf{94.83}\small{$\pm$0.31} \\
      D → E   & 93.66\small{$\pm$0.64} & 93.82\small{$\pm$0.38} & 94.11\small{$\pm$0.34} & 94.31\small{$\pm$0.38}          & 94.45\small{$\pm$0.21}          & \textbf{94.57}\small{$\pm$0.18} \\
      D → K   & 94.16\small{$\pm$0.26} & 94.87\small{$\pm$0.15} & 94.99\small{$\pm$0.25} & 95.54\small{$\pm$0.26}          & 95.36\small{$\pm$0.58}          & \textbf{95.84}\small{$\pm$0.24} \\
      E → B   & 91.58\small{$\pm$0.39} & 92.54\small{$\pm$0.37} & 92.19\small{$\pm$0.66} & 92.99\small{$\pm$0.37}          & \textbf{93.69}\small{$\pm$0.51} & 93.20\small{$\pm$0.63}          \\
      E → D   & 90.32\small{$\pm$0.80} & 91.29\small{$\pm$0.46} & 90.71\small{$\pm$0.69} & 91.38\small{$\pm$0.56}          & 92.35\small{$\pm$0.42}          & \textbf{92.63}\small{$\pm$0.34} \\
      E → K   & 94.73\small{$\pm$0.86} & 95.40\small{$\pm$0.46} & 95.70\small{$\pm$0.58} & 95.76\small{$\pm$0.60}          & \textbf{96.25}\small{$\pm$0.30} & 96.16\small{$\pm$0.07}          \\
      K → B   & 91.85\small{$\pm$0.33} & 91.89\small{$\pm$0.37} & 92.09\small{$\pm$0.53} & \textbf{93.13}\small{$\pm$0.33} & 92.84\small{$\pm$0.47}          & 92.18\small{$\pm$0.84}          \\
      K → D   & 90.32\small{$\pm$0.79} & 89.85\small{$\pm$0.44} & 90.49\small{$\pm$0.70} & 91.15\small{$\pm$0.70}          & \textbf{91.85}\small{$\pm$0.40} & 91.77\small{$\pm$0.68}          \\
      K → E   & 94.07\small{$\pm$0.71} & 94.58\small{$\pm$0.32} & 94.55\small{$\pm$0.26} & 94.85\small{$\pm$0.40}          & 95.00\small{$\pm$0.16}          & \textbf{95.06}\small{$\pm$0.43} \\ \midrule
      AVG     & 92.97\small{$\pm$0.50} & 93.32\small{$\pm$0.38} & 93.46\small{$\pm$0.44} & 93.88\small{$\pm$0.40}          & 94.15\small{$\pm$0.39}          & \textbf{94.17}\small{$\pm$0.40} \\ \midrule
      BDE → K & 95.56\small{$\pm$0.44} & 95.80\small{$\pm$0.37} & 95.84\small{$\pm$0.30} & 96.31\small{$\pm$0.28}          & \textbf{96.46}\small{$\pm$0.26} & 96.13\small{$\pm$0.12}          \\
      BDK → E & 94.94\small{$\pm$0.24} & 95.14\small{$\pm$0.17} & 95.21\small{$\pm$0.27} & 95.25\small{$\pm$0.64}          & 95.50\small{$\pm$0.24}          & \textbf{95.68}\small{$\pm$0.11} \\
      BEK → D & 92.96\small{$\pm$0.30} & 92.95\small{$\pm$0.32} & 93.23\small{$\pm$0.39} & 93.03\small{$\pm$0.10}          & 93.60\small{$\pm$0.26}          & \textbf{93.98}\small{$\pm$0.16} \\
      DEK → B & 93.75\small{$\pm$0.59} & 94.41\small{$\pm$0.46} & 93.92\small{$\pm$0.55} & 94.37\small{$\pm$0.29}          & \textbf{94.77}\small{$\pm$0.35} & 94.57\small{$\pm$0.37}          \\ \midrule
      AVG     & 94.30\small{$\pm$0.39} & 94.58\small{$\pm$0.33} & 94.55\small{$\pm$0.38} & 94.74\small{$\pm$0.33}          & 95.08\small{$\pm$0.28}          & \textbf{95.09}\small{$\pm$0.19} \\ \bottomrule
    \end{tabular}%
  \end{adjustbox}
  \caption{
    Results on Amazon reviews based on RoBERTa$_{base}$.
  }
  \label{roberta-detailed-experiment}
\end{table*}

\end{document}